\documentclass[11pt,a4paper]{article}
\pdfoutput=1

\usepackage[hyperref]{acl2017}
\usepackage{times}
\usepackage{latexsym}

\usepackage{url}

\usepackage{graphicx}
\usepackage{amsmath,amssymb,amsthm}
\usepackage{multirow}
\usepackage{makecell}
\usepackage{url}
\usepackage{color}
\usepackage{caption}
\usepackage{subcaption,siunitx,booktabs}
\usepackage{sidecap}
\usepackage{multirow}
\usepackage{textcomp}
\usepackage{url}
\usepackage{pbox}
\usepackage{dblfloatfix}
\usepackage{here}
 
\usepackage[compact]{titlesec}

\aclfinalcopy 

\renewcommand{\vec}[1]{\mathbf{#1}}
\DeclareMathOperator{\sal}{salience}
 
\title{Graph-based Neural Multi-Document Summarization}
\author{
	    Michihiro Yasunaga\textsuperscript{1} \quad Rui Zhang\textsuperscript{1} \quad Kshitijh Meelu\textsuperscript{1} \\ \quad {\bf Ayush Pareek\textsuperscript{2}} \quad {\bf Krishnan Srinivasan\textsuperscript{1}} \quad {\bf Dragomir Radev\textsuperscript{1}}\\
  	    \textsuperscript{1}Department of Computer Science, Yale University\\
  	    \textsuperscript{2}The LNM Institute of Information Technology\\
        {\small {\tt \{michihiro.yasunaga,r.zhang,kshitijh.meelu\}@yale.edu}}\\[-1mm]
        {\small{\tt \{ayush.original\}@gmail.com}}\\[-1mm]
        {\small{\tt \{krishnan.srinivasan,dragomir.radev\}@yale.edu}}
}

\date{}

\begin{document}
\setlength{\abovedisplayskip}{6pt}
\setlength{\belowdisplayskip}{5pt}
        
\maketitle
\begin{abstract}
We propose a neural multi-document summarization (MDS) system that incorporates sentence relation graphs.
We employ a Graph Convolutional Network (GCN) on the relation graphs, with sentence embeddings obtained from Recurrent Neural Networks as input node features.
Through multiple layer-wise propagation, the GCN generates high-level hidden sentence features for salience estimation.
We then use a greedy heuristic to extract salient sentences while avoiding redundancy.
In our experiments on DUC 2004, we consider three types of sentence relation graphs and demonstrate the advantage of combining sentence relations in graphs with the representation power of deep neural networks.
Our model improves upon traditional graph-based extractive approaches and the vanilla GRU sequence model with no graph, and it achieves competitive results against other state-of-the-art multi-document summarization systems.
\end{abstract}

\section{Introduction}
Document summarization aims to produce fluent and coherent summaries covering salient information in the documents.
Many previous summarization systems employ an extractive approach by identifying and concatenating the most salient text units (often whole sentences) in the document. 
  
Traditional extractive summarizers produce the summary in two steps: sentence ranking and sentence selection.
First, they utilize human-engineered features such as sentence position and length \cite{radev2004mead}, word frequency and importance \cite{nenkova2006compositional,hong2014improving}, among others, to rank sentence salience.
Then, they select summary-worthy sentences using a range of algorithms, such as graph centrality \cite{erkan2004lexrank}, constraint optimization via Integer Linear Programming \cite{mcdonald2007study,gillick2009scalable,li2013using}, or Support Vector Regression \cite{li2007multi} algorithms.
Optionally, sentence reordering \cite{lapata2003probabilistic,barzilay2001sentence} can follow to improve coherence of the summary.

Recently, thanks to their strong representation power, neural approaches have become popular in text summarization, especially in sentence compression \cite{rush2015neural} and single-document summarization \cite{cheng2016neural}.
Despite their popularity, neural networks still have issues when dealing with multi-document summarization (MDS).
In previous neural multi-document summarizers \cite{cao2015ranking,cao2017improving}, 
all the sentences in the same document cluster are processed independently.
Hence, the relationships between sentences and thus the relationships between  different documents are ignored. 
However, \newcite{christensen2013towards} demonstrates the importance of considering discourse relations among sentences in multi-document summarization.
 
This work proposes a multi-document summarization system that exploits the representational power of deep neural networks and the sentence relation information encoded in graph representations of document clusters.
Specifically, we apply Graph Convolutional Networks \cite{kipf2017semi} on sentence relation graphs.
First, we discuss three different
techniques to produce sentence relation graphs, where
nodes represent sentences in a cluster and edges capture the connections between sentences.
Given a relation graph,
our summarization model applies a Graph Convolutional Network (GCN), which takes in sentence embeddings from Recurrent Neural Networks as input node features.
Through multiple layer-wise propagation, the GCN generates high-level hidden features for each sentence that incorporate the graph information.
We then obtain sentence salience estimations via a regression on top, and extract salient sentences in a greedy manner while avoiding redundancy.

We evaluate our model on the DUC 2004 multi-document summarization (MDS) task.
Our model shows a clear advantage over traditional graph-based extractive summarizers, as well as a baseline GRU model that does not use any graph, and achieves competitive results with other state-of-the-art MDS systems.
This work provides a new gateway to incorporating graph-based techniques into neural summarization.
 
\section{Related Work}
\subsection{Graph-based MDS}
Graph-based MDS models have traditionally employed surface level \cite{erkan2004lexrank,mihalcea2005language,wan2006improved} or deep level \cite{pardo2006modeling,antiqueira2009complex} approaches based on topological features and the number of nodes \cite{albert2002statistical}. Efforts have been made to improve decision making of these systems by using discourse relationships between sentences \cite{radev2000common,radev2001newsinessence}.
\newcite{erkan2004lexrank} introduce LexRank to compute sentence importance based on the eigenvector centrality in the connectivity graph of inter-sentence cosine similarity.
\newcite{mei2010divrank} propose DivRank to balance the prestige and diversity of the top ranked vertices in information networks and achieve improved results on MDS.
\newcite{christensen2013towards} build multi-document graphs to identify pairwise ordering constraints over the sentences by accounting for discourse relationships between sentences \cite{mann1988rhetorical}.
In our work, we build on the Approximate Discourse Graph (ADG) model \cite{christensen2013towards} and account for macro level features in sentences to improve sentence salience prediction.

\subsection{Summarization Using Neural Networks}
Neural networks have recently been popular for text summarization \cite{kaageback2014extractive,rush2015neural,yin2015optimizing,cao2016attsum,wang2016neural,cheng2016neural,nallapati2016abstractive,nallapati2017summarunner,see2017get}.
For example, \newcite{rush2015neural} introduce a neural attention feed-forward network-based model for sentence compression.
\newcite{wang2016neural} employ encoder-decoder RNNs to effectively produce short abstractive summaries for opinions.
\newcite{cao2016attsum} develop a query-focused summarization system called AttSum which deals with saliency ranking and relevance ranking using query-attention-weighted CNNs.

Very recently, thanks to large scale news article datasets \cite{hermann2015teaching}, \newcite{cheng2016neural} train an extractive summarization system with attention-based encoder-decoder RNNs to sequentially label summary-worth sentences in single documents.
Moreover, \newcite{see2017get}, adopting an abstractive approach, augment the standard attention-based encoder-decoder RNNs with the ability to copy words from the source text via pointing and to keep track of what has been summarized.
These models \cite{cheng2016neural,see2017get} achieve state-of-the-art performance on single-document summarization tasks.
However, scaling up these RNN sequence-to-sequence approaches to the multi-document summarization task has not been successful, 1) due to the lack of large multi-document summarization datasets needed to train the computationally expensive sequence-to-sequence model, and 2) because of the inadequacy of RNNs to capture the complex discourse relations across multiple documents.
Our multi-document summarization model resolves these issues 1) by 
breaking down the summarization task into salience estimation and sentence selection that do not require an expensive decoder architecture,
and 2) by utilizing sentence relation graphs.

\section{Method}
\begin{figure*}[t]
  \hspace{-2mm}\includegraphics[width=1.02\textwidth]{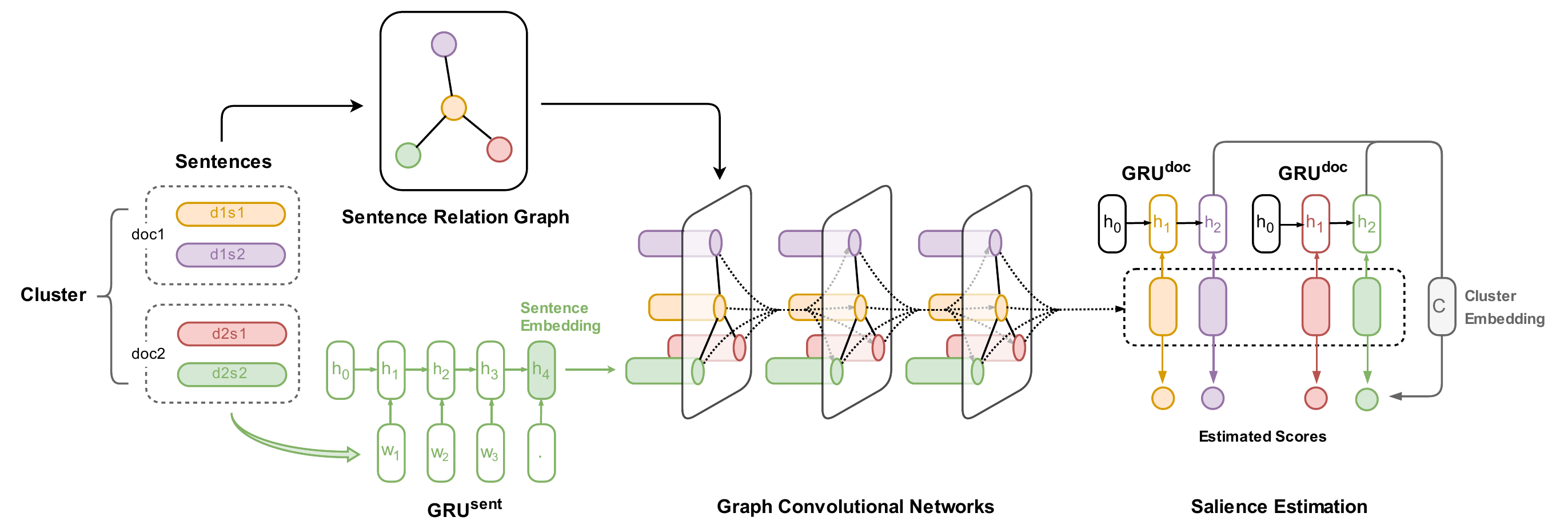}
  \caption{Illustration of our architecture for sentence salience estimation.
  In this example, there are two documents in the cluster and each document has two sentences.
  Sentences are processed by the $\mathrm{GRU}^{sent}$ to get input sentence embeddings.
  The GCN takes the input sentence embeddings and the sentence relation graph, and outputs high-level hidden features for individual sentences.
  $\mathrm{GRU}^{doc}$ produces the cluster embedding from the output sentence embeddings.
  The salience is estimated from the output sentence embeddings and the cluster embedding.
  $w_i$: the word embedding for $i$-th word.
  $h_i$: the hidden state of $\mathrm{GRU}$ at $i$-th step.
  }
  \label{fig:rnn-gcn}
  \vspace{-5mm}
\end{figure*}
Given a document cluster, our method extracts sentences as a summary in two steps: sentence salience estimation and sentence selection.
Figure \ref{fig:rnn-gcn} illustrates our architecture for sentence salience estimation.
Given a document cluster, we first build a sentence relation graph, where interacting sentence nodes are connected by edges. 
For each sentence, we apply an RNN with Gated Recurrent Units ($\mathrm{GRU}^{sent}$) \cite{cho-EtAl:2014:EMNLP2014,chung2014empirical} and extract the last hidden state as the sentence embedding.
We then apply Graph Convolutional Networks \cite{kipf2017semi} on the sentence relation graph with the sentence embeddings as the input node features, to produce final sentence embeddings that reflect the graph representation.
Thereafter, a second level GRU ($\mathrm{GRU}^{doc}$) produces the entire cluster embedding by sequentially connecting the final sentence embeddings.
We estimate the salience of each sentence from the final sentence embeddings and the cluster embedding.
Finally, based on the estimated salience scores, we select sentences in a greedy way until reaching the length limit.

\subsection{Graph Representation of Clusters}

To best evaluate the architecture, we consider three graph representation methods to model sentence relationships within clusters.
First, as prior methods in representing document clusters often adhere to the standard of cosine similarity \cite{erkan2004lexrank}, our initial baseline approach naturally used this representation. Specifically, we add an edge between two sentences if the tf-idf cosine similarity measure between them, using the bag-of-words model,
is above a threshold of 0.2.

\scalebox{0.974}[1.0]{Secondly, the G-Flow system} \cite{christensen2013towards} utilizes discourse relations between sentences to create its graph representations: 
Approximate Discourse Graph (ADG).
The ADG constructs edges between sentences by counting discourse relation indicators such as deverbal noun references, event \!/\! entity continuations, discourse markers, and co-referent mentions. These features allow characterization of sentence relationships, rather than simply their similarity.

While G-Flow's ADG provides many improvements from baseline graph representations, it suffers several disadvantages that diminish its ability to aid
salience prediction when given to the neural network. Specifically, the ADG lacks much diversity in its assigned edge weights. Because the weights are discretely incremented, 
they are multiples of 0.5; many edge weights are 1.0. While the presence of an edge provides a remarkable amount of underlying knowledge on the discourse relationships, edge weights can further include information about the strength 
--- and, similarly, importance --- 
of these relationships. We hope to improve the edge weights by making them more diverse, while infusing more information in the weights themselves. In doing so, we contribute our Personalized Discourse Graph (PDG).
To advance the ADG's performance in providing predictors for sentence salience, we apply a multiplicative effect to the ADG's edge weights via sentence personalization. 

\begin{table}[]
\centering
\begin{tabular}{l}
\multicolumn{1}{c}{Personalization Features} \\\hline
\\[-3.5mm]
\textbullet\hspace{1mm} Position in Document                         \\
\textbullet\hspace{1mm} From $1^{st}$ $3$ Sentences?                   \\
\textbullet\hspace{1mm} No. of Proper Nouns                          \\ 
\textbullet\hspace{1mm} \verb|>| 20 Tokens in Sentence?           \\ 
\textbullet\hspace{1mm} Sentence Length                              \\ 
\textbullet\hspace{1mm} No. of Co-referent Verb Mentions             \\
\textbullet\hspace{1mm} No. of Co-referent Common Noun Mentions      \\
\textbullet\hspace{1mm} No. of Co-referent Proper Noun Mentions      \\

\end{tabular}
\caption{List of features that were input to the regression function in obtaining sentence personalization scores.}
\label{table:personalization}
\vspace{-5mm}
\end{table}

A baseline sentence personalization score $s(v)$, which can be viewed as weighting of sentences, is calculated for every sentence $v$ to account for surface features in each sentence. These features, listed in Table \ref{table:personalization}, are used as input for linear regression, as per \newcite{christensen2013towards}. The regression is applied to each sentence to obtain the personalization score, $s(v)$.
For each sentence, its incoming edge weights in the original ADG are then transformed by the personalization scores and normalized over all the incoming edges.
That is, for directed edge $(u,v) \in E$, the weight is
\begin{equation}
  w_{PDG}(u, v) = \frac{w_{ADG}(u, v) s(u)}{\sum_{u' \in V} w_{ADG}(u', v) s(u')}
\end{equation}
The inclusion of the sentence personalization scores
allows the PDG to account for macro-level features in each sentence, 
augmenting information for salience estimation.
To provide more clarity, we include a figure of the PDG in later sections.

Although it may be possible to incorporate the sentence personalization features later into the salience estimation network, 
we chose to encode them in the PDG to improve the edge weight distribution of sentence relation graphs and to make our salience estimation architecture methodically consistent. 
Additionally, in order to maintain
consistency between  graph  representations, following two modifications  are  made  to  the  discourse  graphs.
First, the directed edges of both the ADG and PDG are made undirected by averaging the edges weights in both directions. Second, edge weights are rescaled to a maximum edge weight of 1 prior to being fed to the GCN.

\subsection{Graph Convolutional Networks}
We apply Graph Convolutional Networks (GCN) from \newcite{kipf2017semi} on top of the sentence relation graph.
In this subsection, we explain in detail the formulation of GCN, and how GCN produces the final sentence embeddings.

The goal of GCN is to learn a function $f(X,A)$ that takes as input:
\begin{itemize}
\setlength{\leftskip}{-2mm}
\setlength{\parskip}{-1mm}
\item $A \in \mathbb{R}^{N \times N}$, the adjacency matrix of graph $\mathcal{G}$, where $N$ is the number of nodes in $\mathcal{G}$.
\item $X \in \mathbb{R}^{N \times D}$, the input node feature matrix, where $D$ is the dimension of input node feature vectors.
\end{itemize}
and outputs high-level hidden features for each node, $Z \in \mathbb{R}^{N \times F}$, that encapsulate the graph structure.
$F$ is the dimension of output feature vectors.
The function $f(X,A)$ takes a form of layer-wise propagation based on neural networks.
We compute the activation matrix in the $(l+1)^{th}$ layer as $H^{(l+1)}$, starting from $H^{0} = X$.
The output of $L$-layer GCN is $Z = f(X,A) = H^{(L)}$.

To introduce the formulation, consider a simple form of layer-wise propagation:
\begin{equation}
\label{eq:simple}
    H^{(l+1)} = \sigma\left(AH^{(l)}W^{(l)}\right)
\end{equation}
where $\sigma$ is an activation function such as $\mathrm{ReLU}(\cdot)$ $= \max(0,\cdot)$.
$W^{(l)}$ is the parameter to learn in the $l^{th}$ layer.
Eq \ref{eq:simple} has two limitations.
First, multiplying by $A$ means that for each node, we sum up the feature vectors of all neighboring nodes but not the node itself.
We fix this by adding self-loops in the graph.
Second, since $A$ is not normalized, multiplying by $A$ will change the scale of feature vectors.
To overcome this, we apply a symmetric normalization by using $D^{-\frac12}AD^{-\frac12}$ where $D$ is the node degree matrix.
These two renormalization tricks result in the following propagation rule:
\begin{equation}
\label{eq:gcn}
    H^{(l+1)} = \sigma\left(\tilde{D}^{-\frac{1}{2}}\tilde{A}\tilde{D}^{-\frac{1}{2}}H^{(l)}W^{(l)}\right)
\end{equation}
where $\tilde{A} = A + I_N$ is the adjacency matrix of the graph $\mathcal{G}$ with added self-loops ($I_N$ is the identity matrix). 
$\tilde{D}$ is the degree matrix with $\tilde{D}_{ii} = \sum_{j}\tilde{A}_{ij}$.
\newcite{kipf2017semi} also provide a theoretical justification of Eq \ref{eq:gcn} as a first-order approximation of spectral graph convolution \cite{hammond2011wavelets,defferrard2016convolutional}.

As an example, if we have a two-layer GCN, we first calculate $\hat{A} = \tilde{D}^{-\frac{1}{2}}\tilde{A}\tilde{D}^{-\frac{1}{2}}$ in a pre-processing step, and then produce
\begin{equation*}
    Z = f(X,A) = \sigma\left(\hat{A}\ \sigma\left(\hat{A}XW^{(0)}\right)W^{(1)}\right)
\end{equation*}

\subsection{Sentence Embeddings}
As the input node features $X$ of GCN,  we use sentence embeddings calculated by $\mathrm{GRU}^{sent}$.

Given a document cluster $C$ with $N$ sentences $(s_1, s_2,..., s_N)$ in total, for each sentence $s_i$ of $L$ words $(w_1, w_2,..., w_L)$, $\mathrm{GRU}^{sent}$ recurrently updates hidden states at each time step $t$:
\begin{equation}
\label{eq:gru_sent}
   \vec{h}_{t}^{sent} = \mathrm{GRU}^{sent}(\vec{h}_{t-1}^{sent},\vec{w}_{t})
\end{equation}
where $\vec{w}_t$ is the word embedding for $w_t$, $\vec{h}_t^{sent}$ is the hidden state of $\mathrm{GRU}^{sent}$.
$\vec{h}_0$ is initialized as a zero vector, and the input sentence embedding $\vec{x}_i$ is the last hidden state:
\begin{equation}
  \vec{x}_i = \vec{h}_L^{sent}
\end{equation}
All sentence embeddings from the given document cluster are grouped as the node feature matrix $X$:
\newcommand*{\vertbar}{\rule[-1ex]{0.5pt}{2.5ex}}
\newcommand*{\horzbar}{\rule[.5ex]{2.5ex}{0.5pt}}
\setlength{\extrarowheight}{1ex}
\begin{equation}
  X =
\left[
  \begin{array}{ccc}
    \horzbar & \vec{x}^{T}_{1} & \horzbar \\
    \horzbar & \vec{x}^{T}_{2} & \horzbar \\
             & \vdots    &          \\
    \horzbar & \vec{x}^{T}_{N} & \horzbar
  \end{array}
\right]  
\end{equation}
$X$ is fed into GCN subsequently to obtain the final sentence embeddings $\vec{s}_{i}$ that incorporate the graph representation of sentence relationships:
\begin{equation}
   Z = f(X,A) =
\left[
  \begin{array}{ccc}
    \horzbar & \vec{s}^{T}_{1} & \horzbar \\
    \horzbar & \vec{s}^{T}_{2} & \horzbar \\
             & \vdots    &          \\
    \horzbar & \vec{s}^{T}_{N} & \horzbar
  \end{array}
\right] 
\end{equation}

\subsection{Cluster Embedding}
Additionally, in order to have a global view of the entire document cluster, we apply a second-level RNN, $\mathrm{GRU}^{doc}$, to encode the entire document cluster.
Given a document cluster $C$ with $M$ documents $(d_1, d_2,..., d_M)$, for document $d_i$ with $|d_i|$ sentences, $\mathrm{GRU}^{doc}$ first builds the document embedding $\vec{d}_i$ on top of sentence embeddings:
\begin{equation}
  \vec{h}_{t}^{doc} = \mathrm{GRU}^{doc}(\vec{h}_{t-1}^{doc},\vec{s}_{t})
\vspace{-5mm}
\end{equation}
\begin{equation}
\label{eq:doc_emb}
  \vec{d}_i = \vec{h}_{|d_i|}^{doc}
\end{equation}
where $\vec{s}_t$ is the sentence embedding in the document $d_i$.
In Eq \ref{eq:doc_emb}, we extract the last hidden state as the document embedding for $d_i$.
In Eq \ref{eq:cluster_emb}, we average over document embeddings to produce the cluster embedding $\vec{C}$:
\begin{equation}
\label{eq:cluster_emb}
  \vec{C} = \frac{1}{M}\sum_{i=1}^{M}\vec{d}_i
\end{equation}
All the GRUs we used are forward.
We also experimented with backward GRUs and bi-directional GRUs, but neither of them meaningfully improved upon forward GRUs. 

\subsection{Salience Estimation}
For the sentence $s_i$ in the cluster $C$, we calculate the salience of $s_i$ as the following, similarly to the attention mechanism in neural machine translation \cite{bahdanau2014neural}:
\begin{equation}
\label{eq:score}
  f(s_i) = \vec{v}^T\tanh(\vec{W}_1 \vec{C} + \vec{W}_2 \vec{s}_i)
\vspace{-5mm}
\end{equation}
\begin{equation}
\label{eq:score-softmax}
  \sal(s_i) = \frac{f(s_i)}{\sum_{s_j \in C}f(s_j)}
\end{equation}
where $\vec{v},\vec{W}_1,\vec{W}_2$ are learnable parameters.
In Eq \ref{eq:score}, we first calculate the score $f(s_i)$ by considering the sentence embedding itself, $\vec{s}_i$, and the cluster embedding $\vec{C}$ for the global context of the multi-document.
The score is then normalized as $\sal(s_i)$ via softmax in Eq \ref{eq:score-softmax}.

\subsection{Training}
The model parameters include the parameters in $\mathrm{GRU}^{sent}$ and $\mathrm{GRU}^{doc}$, the weights in GCN layers, and the parameters for salience estimation ($\vec{v},\vec{W}_1,\vec{W}_2$).
Parameters in $\mathrm{GRU}^{sent}$ and $\mathrm{GRU}^{doc}$ are not shared.
The model is trained end-to-end to minimize the following cross-entropy loss between the salience prediction and the normalized ROUGE score of each sentence:
\begin{equation}
\label{eq:loss}
\mathcal{L} = -\sum_{C}\sum_{s_i \in C} R(s_i)\log(\sal(s_i))
\end{equation}
$R(s_i)$ is calculated by $R(s_i)$ $=$ $\mathrm{softmax}(\alpha\, r(s_i))$, where $r(s_i)$ is
the average of ROUGE-1 and ROUGE-2 Recall scores of sentence $s_i$ by measuring with the ground-truth human-written summaries. 
$\alpha$ is a constant rescaling factor to make the distribution sharper. 
The value of $\alpha$ is determined from the validation data set.
$\alpha r(s_i)$ is then normalized across the cluster via softmax, similarly to Eq \ref{eq:score-softmax}.

\subsection{Sentence Selection}
Given the salience score estimation, we apply a simple greedy procedure to select sentences.
Sentences with higher salience scores have higher priorities.
First, we sort sentences in descending order of the salience scores.
Then, we select one sentence from the top of the list and append to the summary if the sentence is of reasonable length (8-55 words, as in \cite{erkan2004lexrank}) and is not redundant.
The sentence is redundant if the tf-idf cosine similarity between the sentence and the current summary is above 0.5 \cite{hong2014improving}.
We select sentences this way until we reach the length limit.

\section{Experiments}
In this section, we evaluate our model on benchmark MDS data sets, and compare with other state-of-the-art systems.
We aim to show that our model, by combining sentence relations in graphs with the representation power of deep neural networks, can improve upon other traditional graph-based extractive approaches and the vanilla GRU model which does not use any graph.
In addition, we further study the effect of graph and different graph representations on the summarization performance and investigate the correlation of graph structure and sentence salience estimation.

\subsection{Data Set and Evaluation}
We use the benchmark data sets from the Document Understanding Conferences (DUC) containing clusters of English news articles and human reference summaries.
Table \ref{table:duc_dataset} shows the statistics of the data sets.
We use DUC 2001, 2002, 2003 and 2004 containing 30, 59, 30 and 50 clusters of nearly 10 documents each respectively.
Our model is trained on DUC 2001 and 2002, validated on 2003, and tested on 2004.
For evaluation, we use the ROUGE-1,2 metric, with stemming and stop words not removed as suggested by \newcite{owczarzak2012assessment}. 

\begin{table}[t]
\centering
\hspace{-1mm}\scalebox{0.76}{
\begin{tabular}{l|cccc}
                &  DUC'01  &  DUC'02  &  DUC'03  &  DUC'04  \\ \Xhline{4\arrayrulewidth}
\# of Clusters  & 30        & 59        & 30        & 50        \\
\# of Documents & 309       & 567       & 298       & 500       \\
\# of Sentences & 24498     & 16090     & 7721      & 13270     \\
Vocabulary Size & 28188     &  22174    & 13248     & 18036     \\
\shortstack{Summary Length\\~}  & \shortstack{~\vspace{1.5mm}\\100\vspace{-1pt}\\ words} & \shortstack{~\vspace{1.5mm}\\100\vspace{-1pt}\\ words}  & \shortstack{~\vspace{1.5mm}\\100\vspace{-1pt}\\ words} & \shortstack{~\vspace{1.5mm}\\665\vspace{-1pt}\\ Bytes} \\
\end{tabular}
}
\caption{Statistics for DUC Multi-Document Summarization Data Sets.}
\label{table:duc_dataset}
\vspace{-5mm}
\end{table}

\subsection{Experimental Setup}
We conduct four experiments on our model: three using each of the three types of graphs discussed earlier, and one without using any graph. 
In the experiments with graphs, for each document cluster, we tokenize all the documents into sentences and 
generate a graph representation of their relations by the three methods: Cosine Similarity Graph, Approximate Discourse Graph (ADG) from G-Flow, and our Personalized Discourse Graph (PDG). 
Note that for the Cosine Similarity Graph, we compute the tf-idf cosine similarity for every pair of sentences using the bag-of-word model and add an edge for similarity above 0.2. The weight of the edge is the value of similarity.
We apply GCNs with the graphs in the final step of sentence encoding.
For the experiment without any graph, we omit the GCN part and simply use the GRU sentence and cluster encoders.

We use 300-dimensional pre-trained word2vec embeddings \cite{mikolov2013distributed} as input to $\mathrm{GRU}^{sent}$ in Eq \ref{eq:gru_sent}.
The word embeddings are fine-tuned during training.
We use three GCN hidden layers ($L$ = 3).
The hidden states in $\mathrm{GRU}^{sent}$, GCN hidden layers, and $\mathrm{GRU}^{doc}$ are all 300-dimensional vectors ($D=F=300$).

The rescaling factor $\alpha$ in the objective function (Eq \ref{eq:loss}) is chosen as 40 from $\{$10, 20, 30, 40, 50, 100$\}$ based on the validation performance.
The objective function is optimized using Adam \cite{kingma2015adam} stochastic gradient descent with a learning rate of 0.001 and a batch size of 1.
We use gradient clipping with a maximum gradient norm of 1.0.
The model is validated every 10 iterations, and
the training is stopped early if the validation performance does not improve for 10 consecutive steps.
We trained using a single Tesla K80 GPU. 
For all the experiments, the training took approximately 30 minutes until a stop.

\subsection{Results}
\begin{table}[t]
\centering
\hspace{-1.5mm}\scalebox{.73}{
\begin{tabular}{l|cccccc}
          & R-1 & \multicolumn{1}{c}{R-2}  \\ \Xhline{4\arrayrulewidth}
SVR \cite{li2007multi}
              &  36.18	      &  9.34 \\
CLASSY11	  \cite{conroy2011classy}
              & 37.22	      & 9.20  \\
CLASSY04	  \cite{conroy2004left}
              & 37.62	      & 8.96  \\
\scalebox{.95}{GreedyKL	  \cite{haghighi2009exploring}}
              & 37.98	      & 8.53  \\
TsSum	      \cite{conroy2006topic}
              & 35.88	      & 8.15  \\
G-Flow	      \cite{christensen2013towards}
              & 35.30	      & 8.27  \\
FreqSum	    \cite{nenkova2006compositional}
              & 35.30	      & 8.11  \\ 
Centroid	  \cite{radev2004centroid}
              & 36.41	      & 7.97  \\
Cont. LexRank \cite{erkan2004lexrank} 
              & 35.95	      & 7.47  \\
RegSum        \cite{hong2014improving}
              & {\bf 38.57}         & {\bf 9.75}  \\
\hline
GRU & $\text{36.64}_{\pm \text{0.11}}$ & 8.47 \\ 
GRU+GCN: Cosine Similarity Graph & $\text{37.33}_{\pm \text{0.23}}$& 8.78\\
GRU+GCN: ADG from G-Flow &$\text{37.41}_{\pm \text{0.32}}$ & 8.97 \\
GRU+GCN: Personalized Discourse Graph & {\bf $\text{38.23}_{\pm \text{0.22}}$ }& {\bf 9.48}\\
\end{tabular}
}
\caption{ROUGE Recalls on DUC 2004. We show mean (and standard deviation for R-1) over 10 repeated trials for each of our experiments.}
\label{tb:result}
\vspace{-5mm}
\end{table}

Table \ref{tb:result} summarizes our results.
First we take our simple GRU model as the baseline of the RNN-based regression approach.
As seen from the table, the addition of Cosine Similarity Graph on top of the GRU clearly boosts the performance.
Furthermore, the addition of ADG from G-Flow gives a slighly better performance.
Our Personalized Discourse Graph (PDG) enhances the R-1 score by more than 1.50.
The improvement indicates that the combination of graphs and GCNs processes sentence relations across documents better than the vanilla RNN sequence models.

To gain a global view of our performance, we also compare our result with other baseline multi-document summarizers and the state-of-the-art systems related to our regression method.
We compute ROUGE scores from the actual output summary of each system.
We run the G-Flow code released by \newcite{christensen2013towards} to get the output summary of the G-Flow system. The output summary of other systems are compiled in \newcite{hong2014repository}.
To ensure fair comparison, we use ROUGE-1.5.5 with the same parameters as in \newcite{hong2014repository} across all methods: -n 2 -m -l 100 -x -c 95 -r 1000 -f A -p 0.5 -t 0.

From Table \ref{tb:result}, we observe that our GCN system significantly outperforms the commonly used baselines and  traditional
graph approaches such as Centroid, LexRank, and G-Flow.
This indicates the advantage of the representation power of neural networks used in our model.
Our system also exceeds CLASSY04, the best peer system in DUC 2004, and Support Vector Regression (SVR), a widely used regression-based summarizer.
We remain at a comparable level to RegSum, the state-of-the-art multi-document summarizer using regression. 
The major difference is that RegSum performs regression on word level and estimates the salience of each word through a rich set of word features, such as frequency, grammar, context, and hand-crafted dictionaries. 
RegSum then computes sentence salience based on the word scores.
On the other hand, our model simply works on sentence level, spotlighting sentence relations encoded as a graph.
Incorporating more word-level features into our discourse graphs may be an interesting future direction to explore.

\begin{table}[t]
\centering
\hspace{-1mm}\scalebox{0.87}{
\begin{tabular}{l|cccc}
                & \shortstack{PDG\\~}  & \shortstack{ADG\\~}  & \scalebox{0.9}{\shortstack{Cosine \\ Similarity}}   & \scalebox{0.9}{\shortstack{No\\Graph}}  \\
\Xhline{4\arrayrulewidth}
Num of Iterations   & 200        & 280        & 310        & 250    \\
Train Cost          & 4.286      & 5.460      & 5.458      & 5.310  \\
Validation Cost     & 4.559      & 5.077      & 5.099      & 5.214  \\
\end{tabular}
}
\caption{Training statistics for the four experiments.
The first row shows the number of iterations the model took to reach the best validation result before an early stop.
The train cost and validation cost at that time step are shown in the second row and third row, respectively.
All the values are the average over 10 repeated trials.}
\label{tb:iter_and_cost}
\vspace{-5mm}
\end{table}

\begin{figure}[h]
    \hspace{-0.4mm}
    \includegraphics[width=0.47\textwidth]{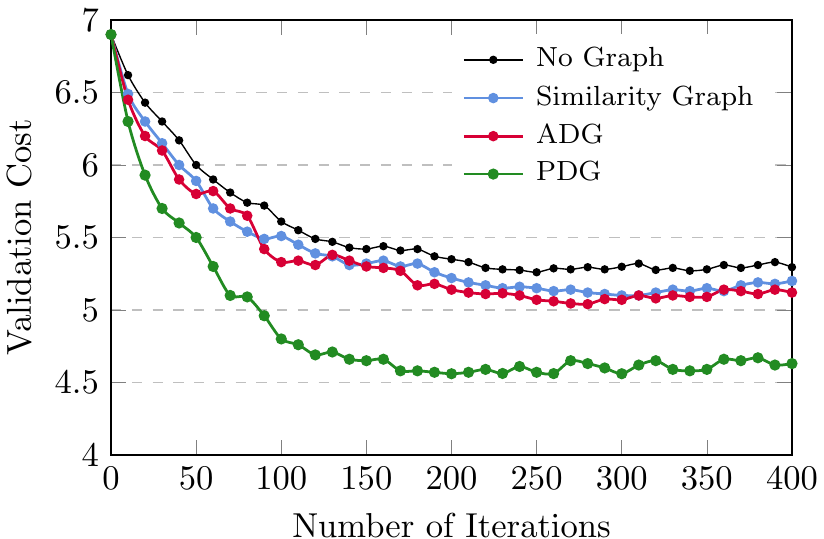}
    \caption{Visualization of the learning curves for the four experiments. The vertical axis displays the validation costs in the interval 4.0 - 7.0.}
    \label{fig:learning_curve}
\end{figure}

\begin{table}[h]
\centering
\hspace{-1.3mm}\scalebox{0.87}{
\begin{tabular}{c|cccccc}
         & \shortstack{PDG\\~}  & \shortstack{ADG\\~}  & \scalebox{0.9}{\shortstack{Cosine \\ Similarity}}  \\ \Xhline{4\arrayrulewidth}
Number of nodes & 265 & 265 & 265 \\
Number of edges & 1023 & 1050 & 884 \\
Average edge weight & 0.075 & 0.295 & 0.359\\
Average node degree & 0.171 & 5.136 & 2.260 \\ \hline
$\rho$ of degree and salience & 0.136 & 0.113 & 0.093\\
\end{tabular}
}
\caption{Characteristics of the three graph representations, averaged over the clusters (i.e. graphs) in DUC 2004. Note that max edge weight in all three representations is 1.0 due to rescaling for consistency. The degree of each node is calculated as the sum of edge weights.}
\label{tb:graph_characteristics}
\vspace{-5mm}
\end{table}

\begin{figure*}[t]
    \hspace{-1mm}\includegraphics[width=1.01\textwidth]{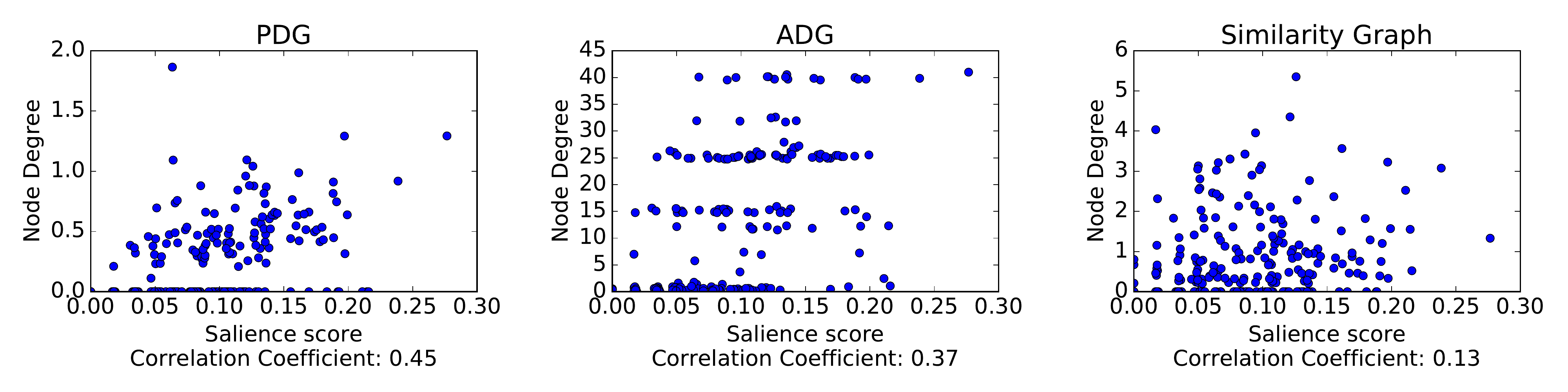}
    \caption{Visualization of the relationship between salience score and node degree for the three graph representation methods. Cluster d30011t from DUC 2004 is chosen as an example.\vspace{3mm}}
    \label{fig:d30011t_scatter}
\vspace{-5mm}
\end{figure*}

\subsection{Discussion}

As shown in Table \ref{tb:result}, our graph-based models outperform the vanilla GRU model, which has no graph.
Additionally, for the three graphs we consider, PDG improves R-1 score by 0.82 over ADG, and ADG outperforms the Cosine Similarity Graph by 0.08 on the R-1 score.
While the Cosine Similarity Graph encodes general word-level connections between sentences, discourse graphs, especially our personalized version, specialize in representing the narrative and logical relations between sentences.
Therefore, we hypothesize that the PDG provides a more informative guide to estimating the importance of each sentence.
In an attempt to better understand the results and validate the effect of sentence relation graphs (especially of the PDG), we have conducted the analysis that follows.

\paragraph{Training Statistics.}
We compare the learning curves of the four different settings: GRU without any graph, GRU+GCN with the Cosine Similarity Graph, GRU+GCN with ADG, and GRU+GCN with PDG (see Table \ref{tb:iter_and_cost} \& Figure \ref{fig:learning_curve}).
Without a graph, the model converges faster and achieves lower training cost than the Cosine Similarity Graph and ADG. This is most likely due to the simplicity of the architecture, but it is also less generalizable, yielding a higher validation cost than the models with graphs.
For the three graph methods, ADG converges faster and has better validation performance than the Cosine Similarity Graph.
PDG converges even faster than ``No Graph" and achieves the lowest training cost and validation cost amongst all methods.
This shows that the PDG has particularly strong representation power and generalizability.

\paragraph{Graph Statistics.}

\begin{figure*}[h]
\centering
    \includegraphics[width=.76\textwidth]{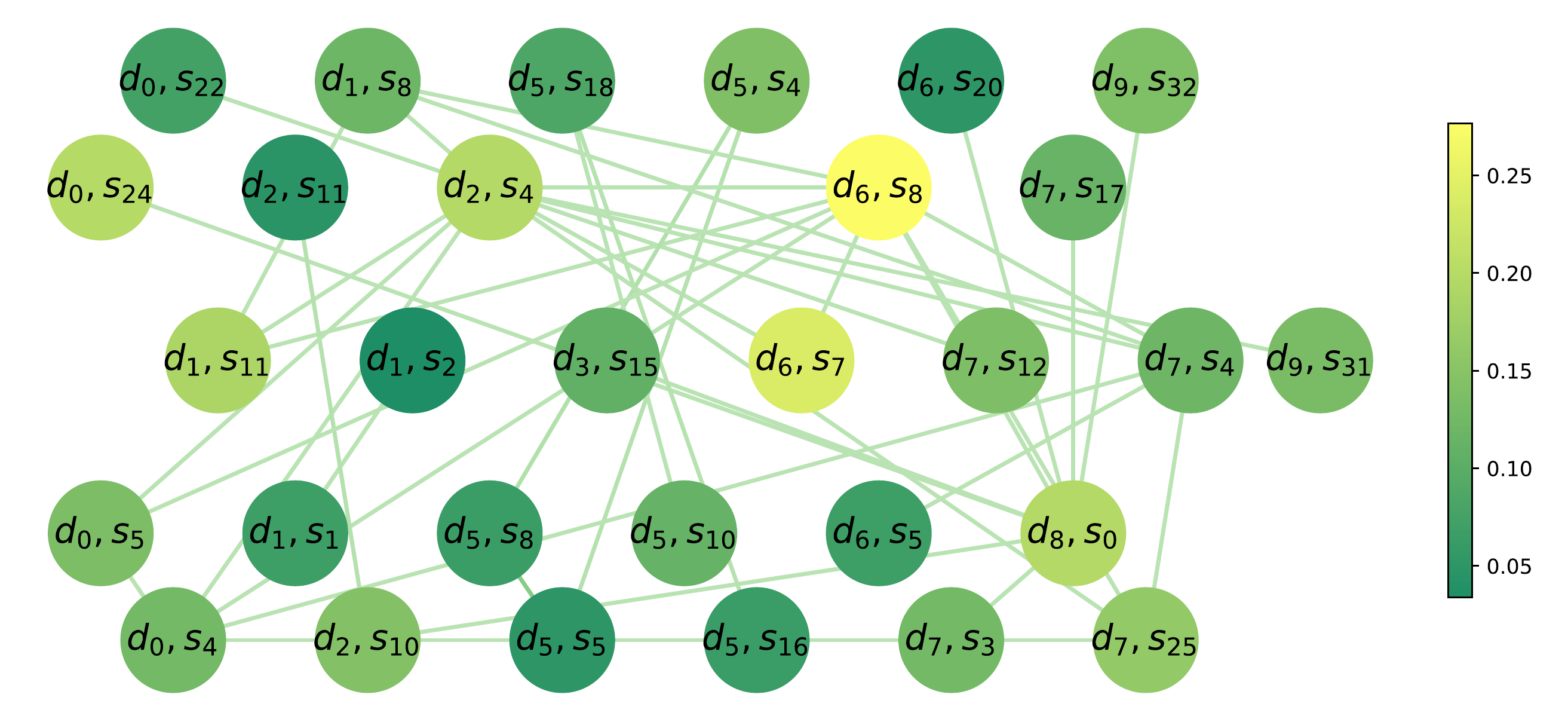}
    \caption{Visualization of the PDG on cluster d30011t. Each node is a sentence, with label (DocumentID, SentenceID). The node color represents the salience score (see the color bar). For simplicity, we only display edges of weight above 0.03. Best viewed in color.
    }
    \label{fig:d30011t_pdg_visual}
\end{figure*}
\begin{figure}[h]
\hspace{2mm}\includegraphics[width=.45\textwidth]{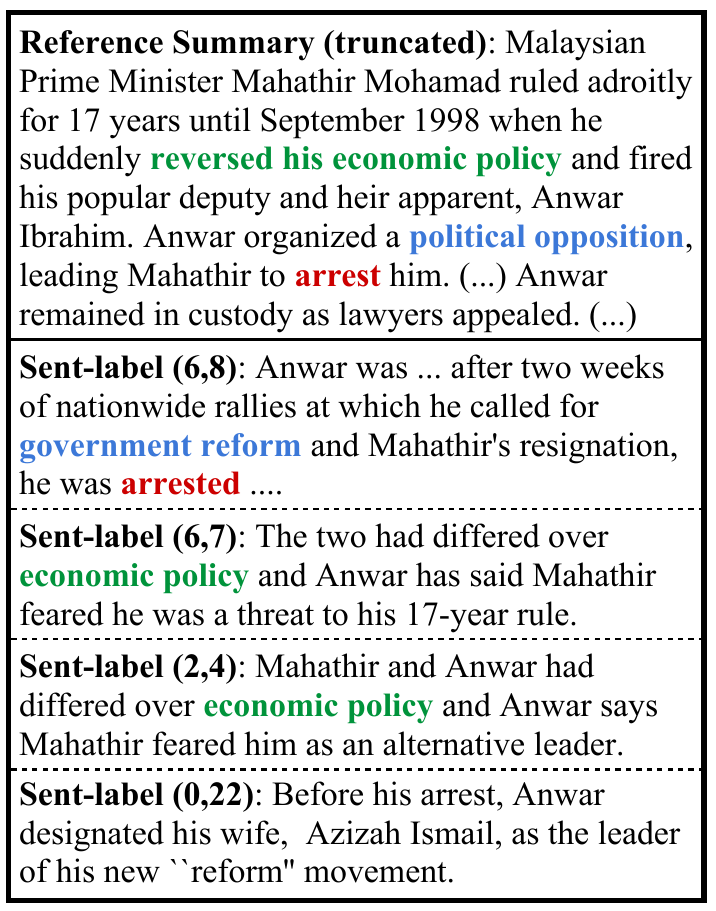}
    \caption{Reference summary and illustrative sentences from cluster d30011t.
    }
    \vspace{-5mm}
    \label{fig:d30011t_reference}
\end{figure}

We also analyze the characteristics of the three graph representation methods on DUC 2004 document clusters.
Table \ref{tb:graph_characteristics} summarizes the following basic statistics: 
the number of nodes (i.e. sentences), the number of edges, average edge weight, and average node degree per graph. We include the correlation between node degree and salience, as well.

As seen from the table, PDG and ADG have approximately the same number of edges. This is expected since the PDG is built by transforming the edge weights in ADG. The Cosine Similarity Graph has slightly fewer edges, simply due to the implemented threshold.

Moreover, note that the ADG has significantly higher average edge weight and node degree as compared to the PDG. These values reflect the discrete nature of the ADG's edge assignment --- further evidence of this can be seen in Figure \ref{fig:d30011t_scatter}. Because the ADG's raw edge weight assignment is done by increments of 0.5, the average node degree tends to be significantly large. This motivated the construction of our PDG, which corrects for this by coercing the average edge weight and node degree to be more diverse and, consequently, smaller (after rescaling).
The process of including sentence personalization scores in edge weight assignments of the PDG leads to a select number of edges gaining markedly large distinction. This aids the GCN in identifying the most important edge connections along with the affiliated sentences.

\paragraph{Node Degree and Salience.}
In Table \ref{tb:graph_characteristics}, we also calculate the correlation coefficient $\rho$, per graph, between the degree of each sentence node and its salience score.
We observe that all the graph representations show positive correlation between the node degree and the salience score.
Moreover, the order of correlation strength is PDG \verb|>| ADG  \verb|>| Cosine Similarity Graph.
Though node degree is a simple measure of these graphs, this observation supports our hypothesis on the efficacy of sentence relation graphs, particularly of PDGs, to provide a guide to salience estimation.
\footnote{
However, we shall add that simply selecting sentences of highest node degrees in PDGs did not itself produce good summaries, compared to our GCN model. Hence, we utilize the graph representations specifically as inputs to the GCN.}

As a case study to illustrate our observation, we chose one cluster (d30011t) from DUC 2004.
Figure \ref{fig:d30011t_scatter} shows the scatter plots of the node degree and salience score of each sentence.

\paragraph{Visualization of the PDG.}
Finally, to demonstrate the functionality of the PDG and complement our discussion from Section 3.1, we visualize the PDG on cluster d30011t with the salience score on each node in Figure \ref{fig:d30011t_pdg_visual} (also see Figure \ref{fig:d30011t_reference} for the actual sentences).

From the visualization, it can be observed that the nodes representing salient sentences (such as ($d_6,s_8$), ($d_6,s_7$), and ($d_2,s_4$)) tend to have higher degrees in the PDG. We can also observe that the PDG represents edges which connect nodes of sentences from different documents, in contrast with the traditional sequence model.

From Figure \ref{fig:d30011t_reference}, we note that the most salient sentence ($d_6,s_8$) actually describes much of the reference summary. As an example of discourse relation,
($d_6,s_7$) and ($d_2,s_4$), the two nodes connected to ($d_6,s_8$), provide the background for ($d_6,s_8$), even though they do not share many words in common with it.
On the other hand,
($d_0,s_{22}$), which is only connected with ($d_2,s_4$), is not salient as it does not provide a central message for the summary.

\section{Conclusion}
In this paper, we \scalebox{0.96}[1]{presented} a novel multi-document summarization system that exploits the representational power of neural networks and graph representations of sentence relationships. 
On top of a simple GRU model as an RNN-based regression baseline, we build a Graph Convolutional Network (GCN) architecture applied on a Personalized Discourse Graph.
Our model, unlike traditional RNN models, 
can capture sentence relations across documents
and demonstrates improved salience prediction and summarization,
achieving competitive performance with current state-of-the-art systems.
Furthermore, through multiple analyses, we have validated the efficacy of sentence relation graphs, particularly of PDG, to help to learn the salience of sentences.
This work shows the promise of the GCN models and of discourse graphs applied to processing multi-document inputs.

\section*{Acknowledgements}
We would like to thank Mirella Lapata, the members of the Sapphire Project (University of Michigan and IBM), as
well as all the anonymous reviewers for their
helpful suggestions on this work. This material is based in part upon work
supported by IBM under contract 4915012629. Any opinions, findings,
conclusions, or recommendations expressed herein are those of the authors and
do not necessarily reflect the views of IBM.

\bibliography{acl2017}
\bibliographystyle{acl_natbib}
\end{document}